\documentclass[10pt,twocolumn,letterpaper]{article}

\usepackage{iccv}              

\usepackage{graphicx}
\usepackage{makecell} 
\usepackage{siunitx}
\usepackage{subcaption}

\definecolor{iccvblue}{rgb}{0.21,0.49,0.74}
\usepackage[pagebackref,breaklinks,colorlinks,allcolors=iccvblue]{hyperref}

\def\paperID{21} 
\def\confName{ICCV}
\def\confYear{2025}

\title{CIPHER: C\textnormal{ulvert} I\textnormal{nspection through} P\textnormal{airwise Frame Selection \\and} H\textnormal{igh-}E\textnormal{fficiency} R\textnormal{econstruction}}

\author{Seoyoung Lee \quad Zhangyang Wang\\
The University of Texas at Austin\\
{\tt\small \{seoyounglee, atlaswang\}@utexas.edu}
}

\begin{document}
\maketitle
\begin{abstract}
Automated culvert inspection systems can help increase the safety and efficiency of flood management operations. As a key step to this system, we present an efficient RGB-based 3D reconstruction pipeline for culvert-like structures in visually repetitive environments. Our approach first selects informative frame pairs to maximize viewpoint diversity while ensuring valid correspondence matching using a plug-and-play module, followed by a reconstruction model that simultaneously estimates RGB appearance, geometry, and semantics in real-time. Experiments demonstrate that our method effectively generates accurate 3D reconstructions and depth maps, enhancing culvert inspection efficiency with minimal human intervention.
\vspace{-6pt}
\end{abstract}
    
\section{Introduction}
\label{sec:intro}

Regular inspections of culvert conditions are essential for preventing catastrophic events and ensuring the safety of flood management systems. Culverts, which are typically circular or box-shaped structures that lie adjacent to or beneath levees and embankments to facilitate water flow, play a critical role in maintaining the safe and timely operation of water control systems~\cite{usace2020}. Defective culverts can cause seepage, erosion, overtopping, or other system failures. 

Manual culvert inspection often involves intensive human resources, costs, and even safety risks. Thus, recent advancements in robotics, sensors, and AI have led to early efforts in developing automated culvert inspection methods~\cite{murphy2011robot, lim2014robotic, guan2014small, gucunski2017rabit, von2015development, ellenberg2016bridge, mueller2017robotic, gibb2018nondestructive, palomer2019inspection, sanchez2019robotic, la2019automated, li2019cable}. However, many of these systems still rely on manual review of inspection outputs (\eg, CCTV footage), requiring significant time and human effort. There remains a critical need for a fully automated solution that can efficiently inspect culverts, and accurately detect and assess structural defects.

A key step in automating culvert inspection involves utilizing visual or other sensory data from dispatched autonomous vehicles to generate high-quality 3D reconstructions of the culvert interior, enabling detailed analysis and identification of potential failure modes. However, many 3D reconstruction models rely on Structure-from-Motion (SfM) techniques~\cite{schoenberger2016sfm, schoenberger2016mvs, Wang_2024_CVPR, leroy2024grounding}, to estimate camera parameters. Unfortunately, SfM struggles in scenarios similar to culvert interiors due to challenges in the input data -- unevenly lit, low-resolution, similar RGB captures from consistent forward-moving dolly shots. These visually repetitive scenes provide limited viewpoint diversity, making feature matching across frames difficult and causing errors that propagate through the reconstruction process. As recent point-based reconstruction models based on NeRF~\cite{mildenhall2020nerf} and 3DGS~\cite{kerbl3Dgaussians} heavily depend on accurate SfM results, they are often unsuitable for these challenging scenarios.
Moreover, though some recent models~\cite{fan2024instantsplat, fan2024large} can perform accurate reconstructions with just a few input views, they often exhibit sensitivity to the choice of inputs. This makes view selection particularly crucial for unconventional scenes, such as culvert interiors, to achieve scene reconstruction.

We propose an efficient 3D reconstruction pipeline for interiors of pipe-like structures, where challenging input conditions, such as unposed RGB video sequences in visually repetitive scenarios, lead to difficulties in feature matching across views. First, a minimal-cost RGB-based frame selection algorithm identifies the most informative image pairs that maximize scene coverage while maintaining sufficient overlap and consistency for valid correspondence matching. This general approach applicable to lightweight, memory-constrained systems, uses relative motion differences of image features, without the need for ground truth camera parameters to enhance reconstruction quality. Second, we apply an RGB-based 3D reconstruction model~\cite{fan2024large} to process these unposed frame pairs into explicit radiance fields, enabling real-time, simultaneous prediction of the scene’s geometry, appearance, and semantics.

\section{Method}
\label{sec:method}

\subsection{Informative Pairwise Frame Selection}

To address the challenges posed by culvert inspection videos, where frames exhibit repetitive views and lack significant viewpoint variation, we introduce a frame selection algorithm to identify the most informative frame pairs for 3D reconstruction. Our goal is to select a pair of frames $(I_a, I_b)$ that maximizes the reconstruction potential by balancing viewpoint diversity and overlap consistency to ensure valid correspondence matching between frames, without relying on ground truth camera poses. 
Two images are chosen for view variation and efficient deployment.
Our selection algorithm follows a multi-step process incorporating feature extraction, motion estimation, and pair evaluation.

Let $\mathcal{I} = \{I_i\}_{i=1}^{N}$ represent the set of input image frames extracted from an RGB video, where $N$ is the total number of frames and $I_i \in \mathbb{R}^{H \times W \times 3}$ is the $i$-th frame with height $H$ and width $W$. 
For each frame $I_i \in \mathcal{I}$, we extract keypoints and descriptors using a feature extractor~\cite{rublee2011orb, lowe2004distinctive, oquab2023dinov2}. Descriptors between pairs of frames are matched based on L2 norm or Hamming distance to output a set of matches $\mathcal{M}_{i,j}$. For a pair of frames, the average motion magnitude $F_{i,j}$ is computed using optical flow, providing an indication of potential viewpoint variation. Furthermore, baseline $\beta_{i,j}$, the spatial distance between matched keypoints, and angular difference $\theta_{i,j}$ from affine transformation between matched keypoints, are calculated.

To select the best frame pair $(I_a, I_b)$, a scoring function is used to evaluate each pair based on motion, baseline, and angular change. The score for each pair is determined by 
\begin{equation}
\text{score}(i, j) = \beta_{i,j} + \left(1 - \frac{\theta_{i,j}}{T_{\text{angle}}}\right)
\end{equation}
where $F_{i,j} \geq T_{\text{flow}}, \, T_{\text{baseline}} \leq \beta_{i,j} \leq \alpha T_{\text{baseline}}, \, \theta_{i,j} \leq T_{\text{angle}}$ for predefined parameters $\alpha$, $T_{\text{flow}}$, $T_{\text{baseline}}$, and $T_{\text{angle}}$.
The pair with the maximum score is selected as the most informative, providing diverse yet consistent views of the culvert-like interior scenes.
\subsection{Semantic 3D Reconstruction}
The selected frame pairs $(I_a, I_b)$ serve as input to LargeSpatialModel (LSM)~\cite{fan2024large}, which reconstructs the culvert environment by estimating geometry, RGB appearance, and semantic content. LSM processes unposed and uncalibrated RGB frames to generate explicit radiance fields, enabling high-speed, efficient inference from just two input images.

LSM first leverages ViT~\cite{dosovitskiy2020image} to predict pixel-aligned point maps that capture 3D geometry. Semantic features from a pre-trained 2D model (LSeg)~\cite{li2022language} are lifted into 3D by projecting the 2D features onto the corresponding 3D point locations in point maps. Geometry and semantic features associated with each 3D point are jointly refined using cross-modal attention to enhance feature consistency. Finally, the parameters of the refined 3D points are regressed by a multi-layer perceptron to output semantic anisotropic 3D Gaussians, which efficiently represent the reconstructed scene into RGB appearance, depth map, and feature fields.

\section{Experiments \& Results}
\label{sec:experiments}

\noindent
\textbf{Datasets.} 
We evaluate our approach on unconstrained, high-resolution in-the-wild culvert inspection videos, sourced both from publicly available online footage and our own data collection efforts on the field. 
Videos were selected based on their visual and structural similarity to the QV-Pipe dataset~\cite{liu2022videopipe}, which was originally curated for video-based defect classification in quick-view urban pipe inspections. Although QV-Pipe is limited in resolution, it contains over 9,000 videos featuring various urban pipe structures and annotated defects. 
As there is a lack of publicly available high-resolution datasets for culvert inspection due to security and privacy protocols, we curated high-resolution online and captured videos with similar content to QV-Pipe videos.

\noindent
\textbf{Pose Estimation and Alignment.} 
The culvert videos only consist of RGB videos and do not include other 3D annotations, such as point clouds, depth maps, or camera intrinsics and extrinsics. Due to the inherent challenges in culvert datasets, we use MASt3R~\cite{leroy2024grounding} with Sim(3) alignment solely based on camera poses, instead of classical SfM pipelines~\cite{schoenberger2016sfm, schoenberger2016mvs} for robust camera pose generation.

\begin{figure}
  \centering
  \begin{subfigure}[b]{0.48\columnwidth}
    \centering
    \includegraphics[trim=150 10 110 25, clip, width=\linewidth]{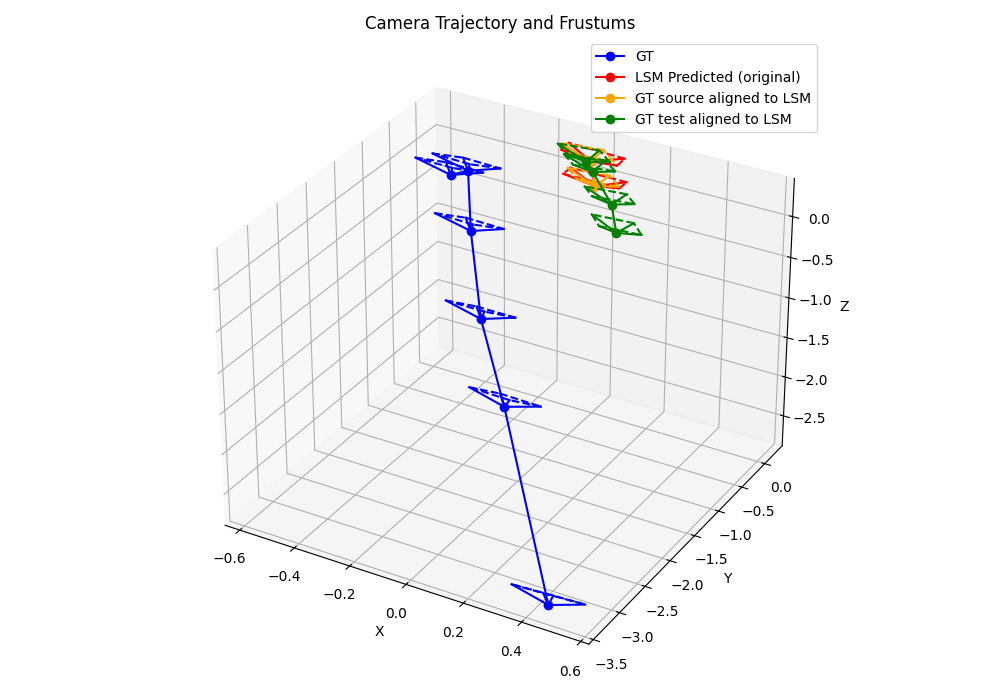}
    \label{fig:img1}
  \end{subfigure}
  \hfill
  \begin{subfigure}[b]{0.48\columnwidth}
    \centering
    \includegraphics[trim=150 10 110 25, clip, width=\linewidth]{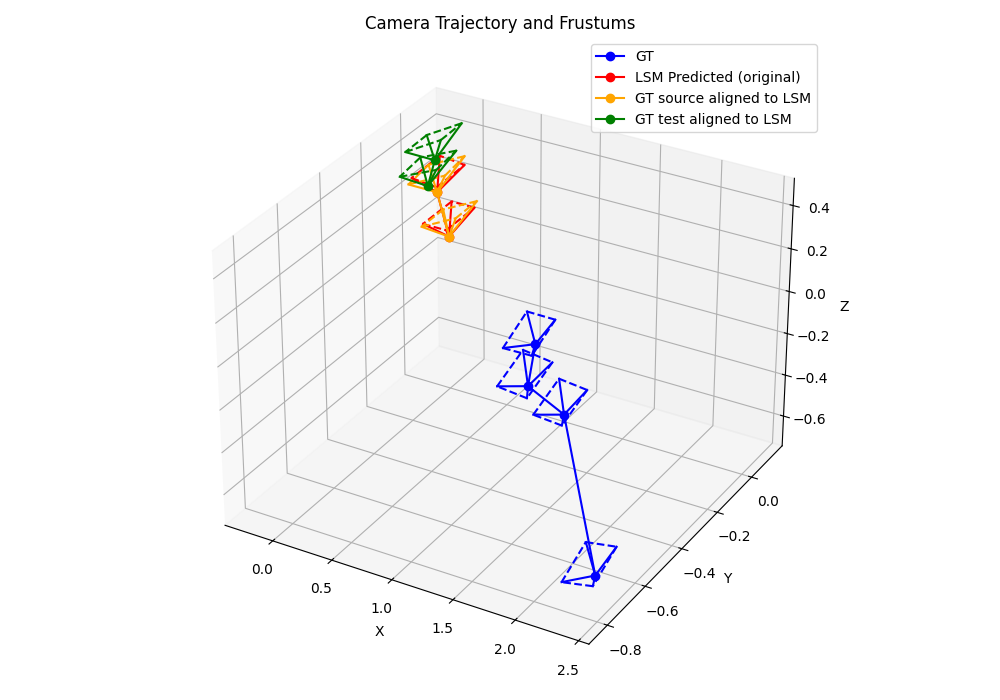}
    \label{fig:img2}
  \end{subfigure}
  \caption{\textbf{Camera pose alignment.} For each scene, ground truth poses were scaled and aligned to match the coordinate system predicted by the reconstruction model.}
  \label{fig:pose_alignment}
\end{figure}

To enable accurate evaluation of rendered views, we align the ground truth test camera poses defined in absolute scale~\cite{leroy2024grounding}, to the coordinate frame of the predicted poses from reconstruction. Since the predicted poses are only determined up to relative coordinates, we estimate a similarity transformation consisting of a scale \( s \), rotation matrix \( \mathbf{R} \in \mathbb{R}^{3 \times 3} \), and translation vector \( \mathbf{t} \in \mathbb{R}^3 \) that best aligns the actual camera centers \( \{\mathbf{C}_i^{\text{GT}}\} \) to the predicted (unscaled) camera centers \( \{\mathbf{C}_i^{\text{pred}}\} \). We compute this similarity transform using Umeyama's method~\cite{umeyama1991least}, a closed-form Procrustes-based algorithm for Sim(3) alignment. The algorithm solves the following minimization problem:
\[
\min_{s, \mathbf{R}, \mathbf{t}} \sum_{i=1}^{N} \left\| s \mathbf{R} \mathbf{C}_i^{\text{GT}} + \mathbf{t} - \mathbf{C}_i^{\text{pred}} \right\|^2.
\]
To align test poses to the predicted coordinate system, we first normalize the ground truth poses and apply the estimated similarity transform \( (s, \mathbf{R}, \mathbf{t}) \) to align them with the predicted poses. 
Unseen poses from these aligned GT poses are used to render novel views at test-time 
to compare reconstruction performance with perceptual metrics such as PSNR, SSIM, and LPIPS. As a result, the two coordinate systems become well aligned as shown in \cref{fig:pose_alignment}.

\noindent
\textbf{Baselines.}
We compare our model against multiple input scenarios and state-of-the-art 3D reconstruction models that have the capability to generate RGB renderings at desired camera poses. Of the RGB renderable models, we prioritized models that are efficient in terms of memory overhead and computation time. We finally selected two models that require SfM to acquire ground truth camera poses to train, of which TensoRF~\cite{Chen2022ECCV} is based on tensorial radiance fields and InstantSplat~\cite{fan2024instantsplat} is based on 3D Gaussian Splatting~\cite{kerbl3Dgaussians}. Specifically, we use the TensoRF implementation in Nerfstudio~\cite{nerfstudio}. InstantSplat is particularly compared for its capability as a sparse-view based model.

\begin{figure} 
\centering
    \includegraphics[width=\linewidth]{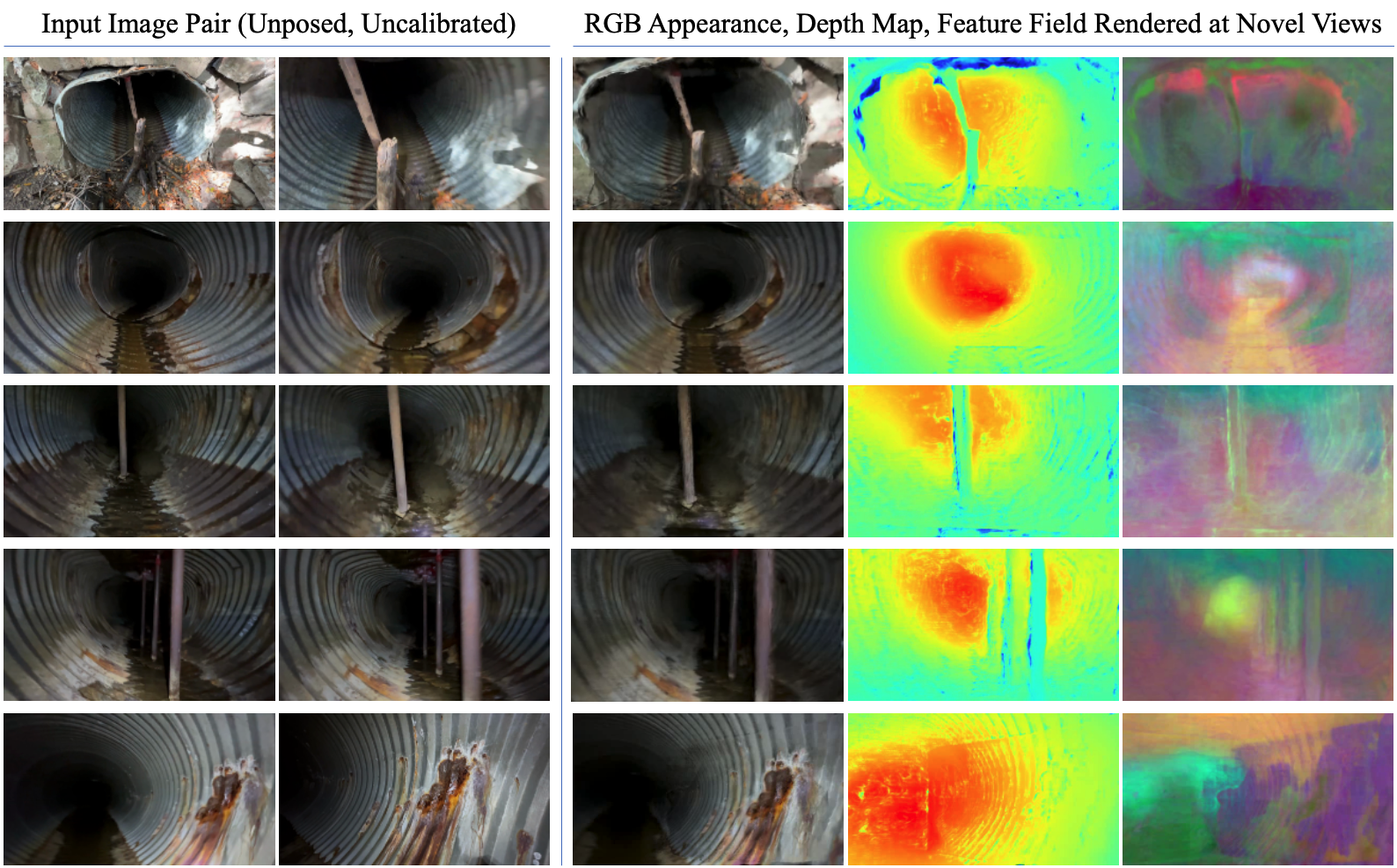} 
\caption{\textbf{Application of pairwise selection algorithm to LSM}. For each scene, two informative frames are selected as input to the reconstruction model, which simultaneously predicts RGB appearance, depth map, and feature field for a smooth interpolated trajectory that includes novel views.} 
\label{fig:results}
\end{figure}

\noindent
\textbf{Evaluation of the Final Approach.}
By applying our approach to culvert inspection videos, we were able to generate view-consistent 3D RGB reconstructions as well as depth maps and feature fields to a certain extent for various structures and defects in culverts. As shown in \cref{fig:results}, the model was able to capture the depth differences inside the culvert even with various lighting conditions and geometric defect structures. It also showed robustness in performance for various camera trajectories, including forward-moving dolly shots, horizontal pans, and diagonal tilts towards the defect (\eg, protruding poles). The semantic information captured via depth maps and feature fields shows that the model can differentiate between different geometric details within the scene, such as the corrosion patterns and pole structures, as well as the depth difference along the longitudinal axis of the culvert.

\begin{figure} 
\centering
    \includegraphics[width=\linewidth]{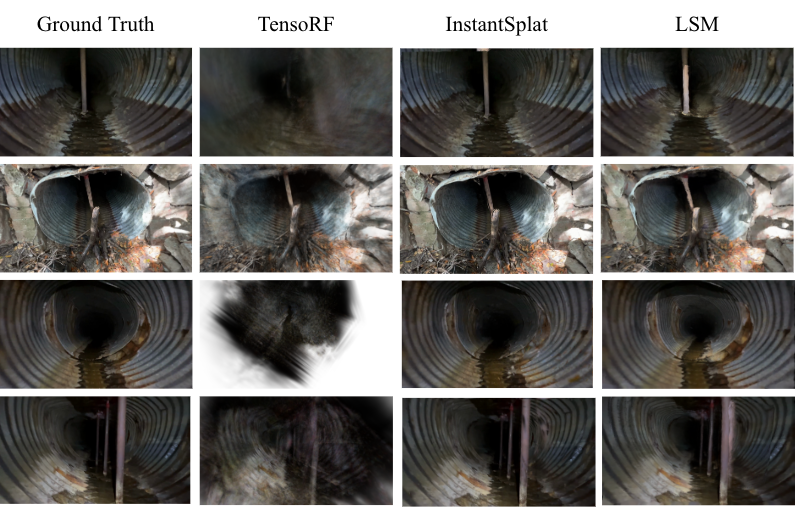} 
\caption{\textbf{Application of pairwise selection algorithm to Baseline models}. The pairwise selection algorithm can help to increase the potential for successful scene reconstruction of novel views for various models in general.} 
\label{fig:baselines}
\end{figure}

We also verified that the high inference efficiency of reconstruction is maintained when our pairwise selection mechanism is applied to LSM. \cref{tab:inference_efficiency} shows performance metrics using 1 NVIDIA RTX A5000 per scene. On average, inference time was less than 1 second per scene. With high resolution outputs, peak GPU memory was around 14 GB, and decreased significantly with lower resolution.

\begin{table}
  \centering
  \small
  \begin{tabular}{@{}lcccc@{}}
    \toprule
    \textbf{\makecell{Image \\ Size}} & \textbf{\makecell{Peak GPU \\ Memory (GB)}} & \textbf{\makecell{Inference \\ Time (sec)}} & \textbf{\makecell{Total Execution \\ Time (sec)}} \\
    \midrule
    $512 \times 288$ & 14.04 & 0.525 & 59.66 \\
    $256 \times 144$ & 8.95 & 0.487 & 48.02 \\
    \bottomrule
  \end{tabular}
  \caption{\textbf{Average inference efficiency of model.} The additional frame selection module shows minimal effect on the efficiency of 3D reconstruction.}
  \label{tab:inference_efficiency}
\end{table}

Compared to other baselines based on tensorial radiance fields and 3D Gaussian Splatting that require camera parameters as input, LSM coupled with our pairwise frame selection algorithm showed the best overall visual quality and accuracy in novel view synthesis, as shown in \cref{fig:baselines}. Moreover, approaches using other models required significantly more computation time, even with 2 input images. While LSM takes less than a minute to fit a model and render novel views, TensoRF takes longer than 30 minutes to train, and InstantSplat can take more than 5 minutes for training to converge for the same input views. As a result, our approach overall showed the best performance in terms of computation efficiency and reconstruction quality.

\noindent
\textbf{Ablation Studies on Frame Selection.}
We conduct ablation studies to demonstrate the effect of using our proposed informative image pair selection algorithm. For each scene in the test dataset, we apply different methods to choose input views to be used in training the 3D reconstruction model. Assuming the input frames are temporally ordered as extracted from a video, we compare random selection, heuristic approaches, and our approach for input selection. As heuristics, we evaluate (1) selecting the first and last frames to maximize view and motion diversity, and (2) selecting frames at the first (Q1) and third (Q3) quartiles of the sequence to achieve a more balanced sampling. Closer intervals were not considered for sufficient scene coverage.

\cref{tab:selection-metrics} presents a comparison of different input image pair selection strategies that can be utilized for training 3D reconstruction models, evaluated using three standard image quality metrics: PSNR, SSIM, and LPIPS. 
Our approach outperforms all baseline strategies by a large margin across all three metrics, especially achieving more than a twofold improvement in PSNR. In contrast, alternative frame selection strategies exhibit substantially lower performances, which we attribute to the frequent reconstruction failures caused by suboptimal frame pair selection. Notably, these baseline strategies led to successful reconstructions in fewer than 50\% of the test cases, whereas it was possible to get reconstruction across all experiment scenarios with our frame selection strategy. These results demonstrate the effectiveness of our informative pair selection in increasing the potential of robust 3D reconstructions that are more accurate and perceptually faithful.

Our frame selection approach also shows promising results as a general strategy to determine the optimal input frames for sparse-view based 3D reconstruction methods. Similar to LSM, InstantSplat also showed increased performance when used with our frame selection method. As shown in the third column of \cref{fig:baselines}, novel views rendered from InstantSplat trained with input views selected with our method can also show comparable results. Renderings from the model trained with views selected from our approach were also robust to view diversity in novel view synthesis, as compared to renderings from models trained on randomly selected input views. Since our method is based on a lightweight heuristic, it incurs minimal computational and memory overhead, making it an easily integrable, model-agnostic add-on for existing reconstruction models.

\begin{table}
\centering
\begin{tabular}{
    l
    S[round-mode=places, round-precision=3, table-format=2.3]
    S[round-mode=places, round-precision=4, table-format=1.4]
    S[round-mode=places, round-precision=4, table-format=1.5]
}
\toprule
\textbf{Selection Strategy} & {\textbf{PSNR ↑}} & {\textbf{SSIM ↑}} & {\textbf{LPIPS ↓}} \\
\midrule
Random Selection    & 5.8412  & 0.0912  & 0.77434 \\
First and Last      & 6.3638  & 0.1156  & 0.73582 \\
Q1 and Q3           & 5.4122  & 0.0504  & 0.81794 \\
\textbf{Ours} & \bfseries 14.124 & \bfseries 0.1924 & \bfseries 0.45808 \\
\bottomrule
\end{tabular}
\caption{Comparison of different frame pair selection strategies for input used to train LSM.}
\label{tab:selection-metrics}
\end{table}

\section{Limitations \& Future Works}
\label{sec:discussion}

Although our approach significantly improves the feasibility of 3D reconstruction in few-view scenarios, it remains susceptible to limitations inherent in the input scenes, such as extreme or inconsistent lighting conditions and the presence of dynamic objects within the culvert. 

In particular, inconsistent illumination inside culverts can adversely affect feature extraction, often resulting in lower-quality reconstructions. 
To address this, we plan to explore the integration of relighting modules, either as a preprocessing step applied to RGB images or as a component within the end-to-end reconstruction pipeline. The goal is to normalize lighting conditions across different sections of the culvert. 
A promising direction is to explicitly model the lighting behavior inside the culvert with a loss term to account for the radiance distribution from a strong, localized light source mounted on the inspection vehicle in an otherwise dark environment.

Moreover, since our objective is to model the static aspects of the scene—namely, the culvert structure and internal defects, we also aim to investigate techniques that provide robust reconstruction even with transient dynamic content. In the input data preprocessing stage, this may involve generative methods for video object inpainting or dynamic element removal~\cite{bian2025videopainteranylengthvideoinpainting, zhang2024avid}, or 3D-aware scene editing approaches~\cite{yang2025storm} that offer more accurate and minimally destructive removal that minimizes effect to the surroundings. 

Ultimately, we aim to incorporate lighting understanding and dynamic content removal into a unified, end-to-end framework for robust 3D reconstruction. To further generalize our informative frame pair selection method beyond the current setting, we plan to explore automated or learnable parameter search for direct applicability to diverse datasets. Moreover, integrating defect detection capabilities into the framework would further enhance its utility for autonomous inspection systems. To support real-time processing and deployment in resource-constrained settings, we also aim to further reduce GPU memory requirements, enabling operation on lightweight embedded systems suited for culvert environments. A key challenge moving forward will be to balance the fidelity of scene reconstruction with the added functionalities, while maintaining computational efficiency.

\section{Conclusion}
\label{sec:conclusion}

This paper presents an efficient RGB-based 3D reconstruction pipeline for the interior of pipe-like structures. To reconstruct challenging inputs, such as unposed, visually similar RGB frames that involve difficulty in feature matching across views, we introduce a low-overhead plug-and-play frame selection algorithm that identifies the most informative input pairs, used to inference RGB appearance, depth, and feature fields in real-time. 
{
    \small
    \bibliographystyle{ieeenat_fullname}
    \bibliography{main}

\begin{thebibliography}{33}
\providecommand{\natexlab}[1]{#1}
\providecommand{\url}[1]{\texttt{#1}}
\expandafter\ifx\csname urlstyle\endcsname\relax
  \providecommand{\doi}[1]{doi: #1}\else
  \providecommand{\doi}{doi: \begingroup \urlstyle{rm}\Url}\fi

\bibitem[Bian et~al.(2025)Bian, Zhang, Ju, Cao, Xie, Shan, and Xu]{bian2025videopainteranylengthvideoinpainting}
Yuxuan Bian, Zhaoyang Zhang, Xuan Ju, Mingdeng Cao, Liangbin Xie, Ying Shan, and Qiang Xu.
\newblock Videopainter: Any-length video inpainting and editing with plug-and-play context control, 2025.

\bibitem[Chen et~al.(2022)Chen, Xu, Geiger, Yu, and Su]{Chen2022ECCV}
Anpei Chen, Zexiang Xu, Andreas Geiger, Jingyi Yu, and Hao Su.
\newblock Tensorf: Tensorial radiance fields.
\newblock In \emph{European Conference on Computer Vision (ECCV)}, 2022.

\bibitem[Dosovitskiy et~al.(2020)Dosovitskiy, Beyer, Kolesnikov, Weissenborn, Zhai, Unterthiner, Dehghani, Minderer, Heigold, Gelly, et~al.]{dosovitskiy2020image}
Alexey Dosovitskiy, Lucas Beyer, Alexander Kolesnikov, Dirk Weissenborn, Xiaohua Zhai, Thomas Unterthiner, Mostafa Dehghani, Matthias Minderer, Georg Heigold, Sylvain Gelly, et~al.
\newblock An image is worth 16x16 words: Transformers for image recognition at scale.
\newblock \emph{arXiv preprint arXiv:2010.11929}, 2020.

\bibitem[Ellenberg et~al.(2016)Ellenberg, Kontsos, Moon, and Bartoli]{ellenberg2016bridge}
Andrew Ellenberg, Ahsas Kontsos, Franklin Moon, and Ivan Bartoli.
\newblock Bridge deck delamination identification from unmanned aerial vehicle infrared imagery.
\newblock \emph{Automation in Construction}, 72:\penalty0 155--165, 2016.

\bibitem[Fan et~al.(2024{\natexlab{a}})Fan, Wen, Cong, Wang, Zhang, Ding, Xu, Ivanovic, Pavone, Pavlakos, Wang, and Wang]{fan2024instantsplat}
Zhiwen Fan, Kairun Wen, Wenyan Cong, Kevin Wang, Jian Zhang, Xinghao Ding, Danfei Xu, Boris Ivanovic, Marco Pavone, Georgios Pavlakos, Zhangyang Wang, and Yue Wang.
\newblock Instantsplat: Sparse-view gaussian splatting in seconds, 2024{\natexlab{a}}.

\bibitem[Fan et~al.(2024{\natexlab{b}})Fan, Zhang, Cong, Wang, Li, Wen, Zhou, Kadambi, Wang, Xu, et~al.]{fan2024large}
Zhiwen Fan, Jian Zhang, Wenyan Cong, Peihao Wang, Renjie Li, Kairun Wen, Shijie Zhou, Achuta Kadambi, Zhangyang Wang, Danfei Xu, et~al.
\newblock Large spatial model: End-to-end unposed images to semantic 3d.
\newblock \emph{Advances in Neural Information Processing Systems}, 37:\penalty0 40212--40229, 2024{\natexlab{b}}.

\bibitem[Gibb et~al.(2018)Gibb, La, Le, Nguyen, Schmid, and Pham]{gibb2018nondestructive}
Spencer Gibb, Hung~Manh La, Tuan Le, Luan Nguyen, Ryan Schmid, and Huy Pham.
\newblock Nondestructive evaluation sensor fusion with autonomous robotic system for civil infrastructure inspection.
\newblock \emph{Journal of Field Robotics}, 35\penalty0 (6):\penalty0 988--1004, 2018.

\bibitem[Guan et~al.(2014)Guan, Yan, Yang, and Xu]{guan2014small}
Da Guan, Lei Yan, Yibo Yang, and Wenfu Xu.
\newblock A small climbing robot for the intelligent inspection of nuclear power plants.
\newblock In \emph{2014 4th IEEE international conference on information science and technology}, pages 484--487. IEEE, 2014.

\bibitem[Gucunski et~al.(2017)Gucunski, Basily, Kim, Yi, Duong, Dinh, Kee, and Maher]{gucunski2017rabit}
Nenad Gucunski, Basily Basily, Jinyoung Kim, Jingang Yi, Trung Duong, Kien Dinh, Seong-Hoon Kee, and Ali Maher.
\newblock Rabit: Implementation, performance validation and integration with other robotic platforms for improved management of bridge decks.
\newblock \emph{International Journal of Intelligent Robotics and Applications}, 1:\penalty0 271--286, 2017.

\bibitem[Kerbl et~al.(2023)Kerbl, Kopanas, Leimk{\"u}hler, and Drettakis]{kerbl3Dgaussians}
Bernhard Kerbl, Georgios Kopanas, Thomas Leimk{\"u}hler, and George Drettakis.
\newblock 3d gaussian splatting for real-time radiance field rendering.
\newblock \emph{ACM Transactions on Graphics}, 42\penalty0 (4), 2023.

\bibitem[La et~al.(2019)La, Dinh, Pham, Ha, and Pham]{la2019automated}
Hung~Manh La, Tran~Hiep Dinh, Nhan~Huu Pham, Quang~Phuc Ha, and Anh~Quyen Pham.
\newblock Automated robotic monitoring and inspection of steel structures and bridges.
\newblock \emph{Robotica}, 37\penalty0 (5):\penalty0 947--967, 2019.

\bibitem[Leroy et~al.(2024)Leroy, Cabon, and Revaud]{leroy2024grounding}
Vincent Leroy, Yohann Cabon, and J{\'e}r{\^o}me Revaud.
\newblock Grounding image matching in 3d with mast3r.
\newblock In \emph{European Conference on Computer Vision}, pages 71--91. Springer, 2024.

\bibitem[Li et~al.(2022)Li, Weinberger, Belongie, Koltun, and Ranftl]{li2022language}
Boyi Li, Kilian~Q Weinberger, Serge Belongie, Vladlen Koltun, and Ren{\'e} Ranftl.
\newblock Language-driven semantic segmentation.
\newblock \emph{arXiv preprint arXiv:2201.03546}, 2022.

\bibitem[Li et~al.(2019)Li, Gao, Guo, He, and Shao]{li2019cable}
Xinke Li, Chao Gao, Yongcai Guo, Fuliang He, and Yanhua Shao.
\newblock Cable surface damage detection in cable-stayed bridges using optical techniques and image mosaicking.
\newblock \emph{Optics \& Laser Technology}, 110:\penalty0 36--43, 2019.

\bibitem[Lim et~al.(2014)Lim, La, and Sheng]{lim2014robotic}
Ronny~Salim Lim, Hung~Manh La, and Weihua Sheng.
\newblock A robotic crack inspection and mapping system for bridge deck maintenance.
\newblock \emph{IEEE Transactions on Automation Science and Engineering}, 11\penalty0 (2):\penalty0 367--378, 2014.

\bibitem[Liu et~al.(2022)Liu, Zhang, Li, Liang, Jiang, Qiu, Tang, Xie, Yao, Dai, et~al.]{liu2022videopipe}
Yi Liu, Xuan Zhang, Ying Li, Guixin Liang, Yabing Jiang, Lixia Qiu, Haiping Tang, Fei Xie, Wei Yao, Yi Dai, et~al.
\newblock Videopipe 2022 challenge: Real-world video understanding for urban pipe inspection.
\newblock In \emph{2022 26th International Conference on Pattern Recognition (ICPR)}, pages 4967--4973. IEEE, 2022.

\bibitem[Lowe(2004)]{lowe2004distinctive}
David~G Lowe.
\newblock Distinctive image features from scale-invariant keypoints.
\newblock \emph{International journal of computer vision}, 60:\penalty0 91--110, 2004.

\bibitem[Mildenhall et~al.(2020)Mildenhall, Srinivasan, Tancik, Barron, Ramamoorthi, and Ng]{mildenhall2020nerf}
Ben Mildenhall, Pratul~P. Srinivasan, Matthew Tancik, Jonathan~T. Barron, Ravi Ramamoorthi, and Ren Ng.
\newblock Nerf: Representing scenes as neural radiance fields for view synthesis.
\newblock In \emph{ECCV}, 2020.

\bibitem[Mueller et~al.(2017)Mueller, Fromm, Buelow, Birk, Garsch, and Gebbeken]{mueller2017robotic}
Christian~A Mueller, Tobias Fromm, Heiko Buelow, Andreas Birk, Maximilian Garsch, and Norbert Gebbeken.
\newblock Robotic bridge inspection within strategic flood evacuation planning.
\newblock In \emph{OCEANS 2017-Aberdeen}, pages 1--6. IEEE, 2017.

\bibitem[Murphy et~al.(2011)Murphy, Steimle, Hall, Lindemuth, Trejo, Hurlebaus, Medina-Cetina, and Slocum]{murphy2011robot}
Robin~R Murphy, Eric Steimle, Michael Hall, Michael Lindemuth, David Trejo, Stefan Hurlebaus, Zenon Medina-Cetina, and Daryl Slocum.
\newblock Robot-assisted bridge inspection.
\newblock \emph{Journal of Intelligent \& Robotic Systems}, 64\penalty0 (1):\penalty0 77--95, 2011.

\bibitem[Oquab et~al.(2023)Oquab, Darcet, Moutakanni, Vo, Szafraniec, Khalidov, Fernandez, Haziza, Massa, El-Nouby, et~al.]{oquab2023dinov2}
Maxime Oquab, Timoth{\'e}e Darcet, Th{\'e}o Moutakanni, Huy Vo, Marc Szafraniec, Vasil Khalidov, Pierre Fernandez, Daniel Haziza, Francisco Massa, Alaaeldin El-Nouby, et~al.
\newblock Dinov2: Learning robust visual features without supervision.
\newblock \emph{arXiv preprint arXiv:2304.07193}, 2023.

\bibitem[Palomer et~al.(2019)Palomer, Ridao, and Ribas]{palomer2019inspection}
Albert Palomer, Pere Ridao, and David Ribas.
\newblock Inspection of an underwater structure using point-cloud slam with an auv and a laser scanner.
\newblock \emph{Journal of field robotics}, 36\penalty0 (8):\penalty0 1333--1344, 2019.

\bibitem[Rublee et~al.(2011)Rublee, Rabaud, Konolige, and Bradski]{rublee2011orb}
Ethan Rublee, Vincent Rabaud, Kurt Konolige, and Gary Bradski.
\newblock Orb: An efficient alternative to sift or surf.
\newblock In \emph{2011 International conference on computer vision}, pages 2564--2571. Ieee, 2011.

\bibitem[Sanchez-Cuevas et~al.(2019)Sanchez-Cuevas, Ramon-Soria, Arrue, Ollero, and Heredia]{sanchez2019robotic}
Pedro~J Sanchez-Cuevas, Pablo Ramon-Soria, Bego{\~n}a Arrue, Anibal Ollero, and Guillermo Heredia.
\newblock Robotic system for inspection by contact of bridge beams using uavs.
\newblock \emph{Sensors}, 19\penalty0 (2):\penalty0 305, 2019.

\bibitem[Sch\"{o}nberger and Frahm(2016)]{schoenberger2016sfm}
Johannes~Lutz Sch\"{o}nberger and Jan-Michael Frahm.
\newblock {Structure-from-Motion Revisited}.
\newblock In \emph{Conference on Computer Vision and Pattern Recognition (CVPR)}, 2016.

\bibitem[Sch\"{o}nberger et~al.(2016)Sch\"{o}nberger, Zheng, Pollefeys, and Frahm]{schoenberger2016mvs}
Johannes~Lutz Sch\"{o}nberger, Enliang Zheng, Marc Pollefeys, and Jan-Michael Frahm.
\newblock {Pixelwise View Selection for Unstructured Multi-View Stereo}.
\newblock In \emph{European Conference on Computer Vision (ECCV)}, 2016.

\bibitem[Tancik et~al.(2023)Tancik, Weber, Ng, Li, Yi, Kerr, Wang, Kristoffersen, Austin, Salahi, Ahuja, McAllister, and Kanazawa]{nerfstudio}
Matthew Tancik, Ethan Weber, Evonne Ng, Ruilong Li, Brent Yi, Justin Kerr, Terrance Wang, Alexander Kristoffersen, Jake Austin, Kamyar Salahi, Abhik Ahuja, David McAllister, and Angjoo Kanazawa.
\newblock Nerfstudio: A modular framework for neural radiance field development.
\newblock In \emph{ACM SIGGRAPH 2023 Conference Proceedings}, 2023.

\bibitem[Umeyama(1991)]{umeyama1991least}
Shinji Umeyama.
\newblock Least-squares estimation of transformation parameters between two point patterns.
\newblock \emph{IEEE Transactions on Pattern Analysis \& Machine Intelligence}, 13\penalty0 (04):\penalty0 376--380, 1991.

\bibitem[{U.S. Army Corps of Engineers (USACE)}(2020)]{usace2020}
{U.S. Army Corps of Engineers (USACE)}.
\newblock \emph{{Conduits, Pipes, and Culverts Associated with Dams and Levee Systems EM 1110-2-2902}}, 2020.

\bibitem[von Ellenrieder(2015)]{von2015development}
Karl~D von Ellenrieder.
\newblock Development of a usv-based bridge inspection system.
\newblock In \emph{OCEANS 2015-MTS/IEEE Washington}, pages 1--10. IEEE, 2015.

\bibitem[Wang et~al.(2024)Wang, Leroy, Cabon, Chidlovskii, and Revaud]{Wang_2024_CVPR}
Shuzhe Wang, Vincent Leroy, Yohann Cabon, Boris Chidlovskii, and Jerome Revaud.
\newblock Dust3r: Geometric 3d vision made easy.
\newblock In \emph{Proceedings of the IEEE/CVF Conference on Computer Vision and Pattern Recognition (CVPR)}, pages 20697--20709, 2024.

\bibitem[Yang et~al.(2025)Yang, Huang, Chen, Wang, Li, You, Igl, Sharma, Karkus, Xu, Ivanovic, Wang, and Pavone]{yang2025storm}
Jiawei Yang, Jiahui Huang, Yuxiao Chen, Yan Wang, Boyi Li, Yurong You, Maximilian Igl, Apoorva Sharma, Peter Karkus, Danfei Xu, Boris Ivanovic, Yue Wang, and Marco Pavone.
\newblock Storm: Spatio-temporal reconstruction model for large-scale outdoor scenes.
\newblock \emph{arXiv preprint arXiv:2501.00602}, 2025.

\bibitem[Zhang et~al.(2024)Zhang, Wu, Wang, Luo, Zhang, Zhao, Vajda, Metaxas, and Yu]{zhang2024avid}
Zhixing Zhang, Bichen Wu, Xiaoyan Wang, Yaqiao Luo, Luxin Zhang, Yinan Zhao, Peter Vajda, Dimitris Metaxas, and Licheng Yu.
\newblock Avid: Any-length video inpainting with diffusion model.
\newblock In \emph{Proceedings of the IEEE/CVF Conference on Computer Vision and Pattern Recognition}, pages 7162--7172, 2024.

\end{thebibliography}
}

\end{document}